\pdfoutput=1
\documentclass[11pt]{article}

\usepackage{acl}
\usepackage{microtype}
\usepackage{latexsym}

\usepackage{kotex}
\usepackage{times}
\usepackage{soul}
\usepackage{url}
\usepackage{subfig}
\usepackage{caption}
\usepackage{graphics}
\usepackage[utf8]{inputenc}
\usepackage{graphicx}
\usepackage{amsmath}
\usepackage{amsthm}
\usepackage{booktabs}
\usepackage{algorithm}
\usepackage{algorithmic}
\usepackage{booktabs}
\usepackage{multirow}
\usepackage{adjustbox}
\usepackage{float}
\usepackage{tablefootnote}
\usepackage[T1]{fontenc}
\usepackage[utf8]{inputenc}

\title{Don't Judge a Language Model by Its Last Layer: Contrastive Learning with Layer-Wise Attention Pooling}

\author{
  Dongsuk Oh\footnote[1]{}\,\;\footnote[3]{}, Yejin Kim\footnote[1]{}\,\;\footnote[4]{}, Hodong Lee\footnote[3]{}, H. Howie Huang\footnote[2]{}\,\;\footnote[4]{}\;  and Heuiseok Lim\footnote[2]{}\,\;\footnote[3]{} \\
  \footnote[3]{}\,\;Computer Science and Engineering, Korea University \\
  \footnote[4]{}\,\;Graph Lab., George Washington University \\
  \texttt{\{inow3555,bigshane319,limhseok\}@korea.ac.kr}\\
  \texttt{\{yejinjenny, howie\}@gwu.edu}\\
  }

\begin{document}
\maketitle

\footnotetext{\footnote[1]{} These authors contributed equally.}
\footnotetext{\footnote[2]{} These authors are corresponding authors.}

\begin{abstract}
Recent pre-trained language models (PLMs) achieved great success on many natural language processing tasks through learning linguistic features and contextualized sentence representation. Since attributes captured in stacked layers of PLMs are not clearly identified, straightforward approaches such as embedding the last layer are commonly preferred to derive sentence representations from PLMs. This paper introduces the attention-based pooling strategy, which enables the model to preserve layer-wise signals captured in each layer and learn digested linguistic features for downstream tasks. The contrastive learning objective can adapt the layer-wise attention pooling to both unsupervised and supervised manners. It results in regularizing the anisotropic space of pre-trained embeddings and being more uniform. We evaluate our model on standard semantic textual similarity (STS) and semantic search tasks. As a result, our method improved the performance of the base contrastive learned BERT$_{base}$ and variants.
\end{abstract}

\section{Introduction}
Pre-trained language models (PLMs) \citep{kenton2019bert, liu2019roberta, radford2019language, raffel2019exploring} have shown competitive performance on many natural language processing (NLP) tasks. Also, contrastive learning using the PLMs shows the highest performance in sentence representation. Contrastive learning is to learn effective representations by staying semantically close sample pairs together while dissimilar ones are far apart\cite{hadsell2006dimensionality}.

In general, PLMs use either $[CLS]$ tokens in the last layer, $AVG$ which is the average representation of tokens in the last layer\cite{reimers2019sentence, li2020sentence}, or $AVG_{FL}$ which is the average representation of tokens in the first and last layers\cite{gao2021simcse}, to pool out sentence representation from word representations. However, since language models show performance gaps by domain when trained on different objectives, the fixed pooling strategy has limitations in performance improvement. 

\begin{figure}[t]
\centering
	\includegraphics[width=0.9\linewidth]{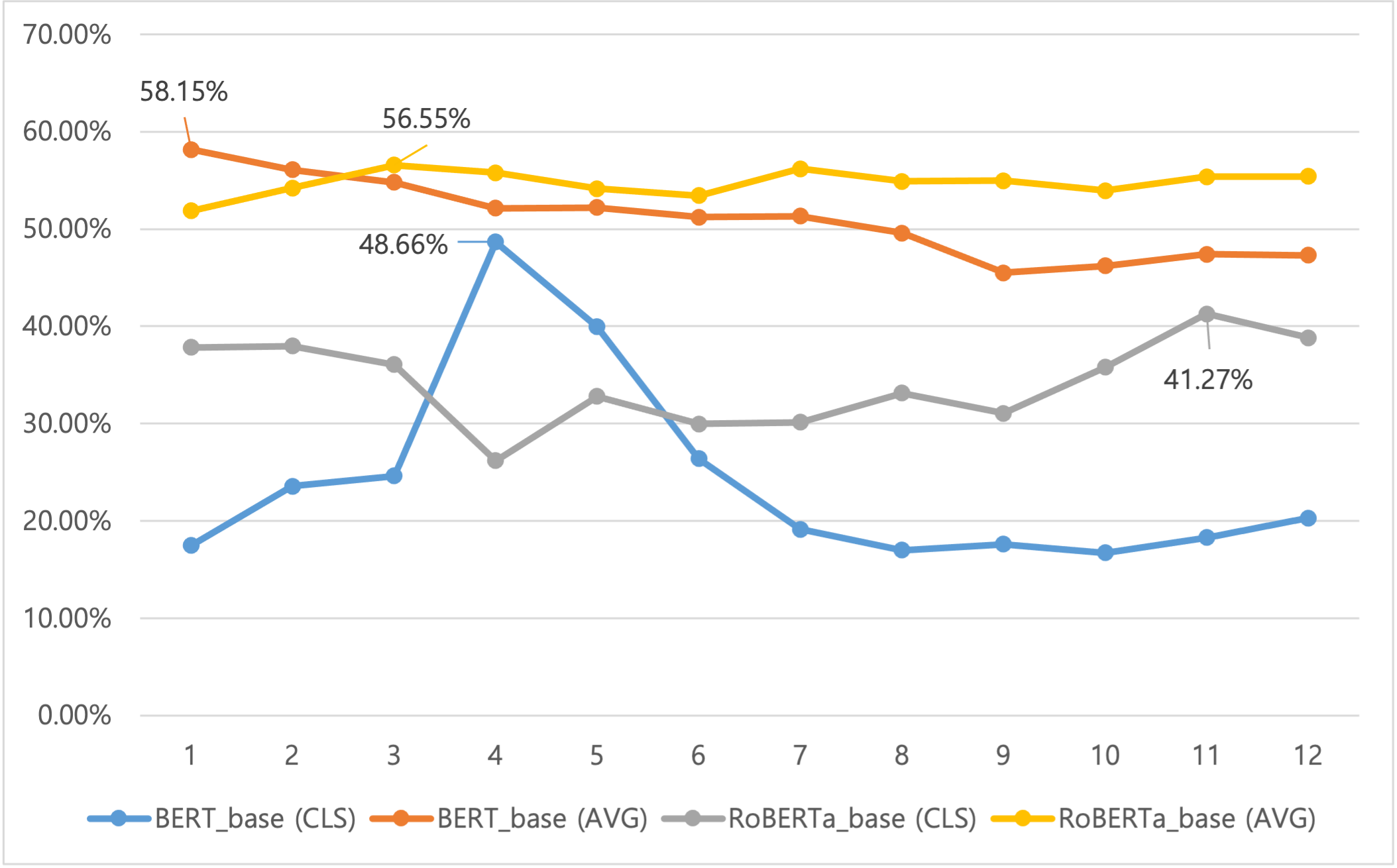}
\caption{Spearsman's correlation score of each layer evaluated on STS-B test set}
\label{fig:intro_fig1}
\end{figure}

\begin{figure}[t]
\centering
	\includegraphics[width=0.9\linewidth]{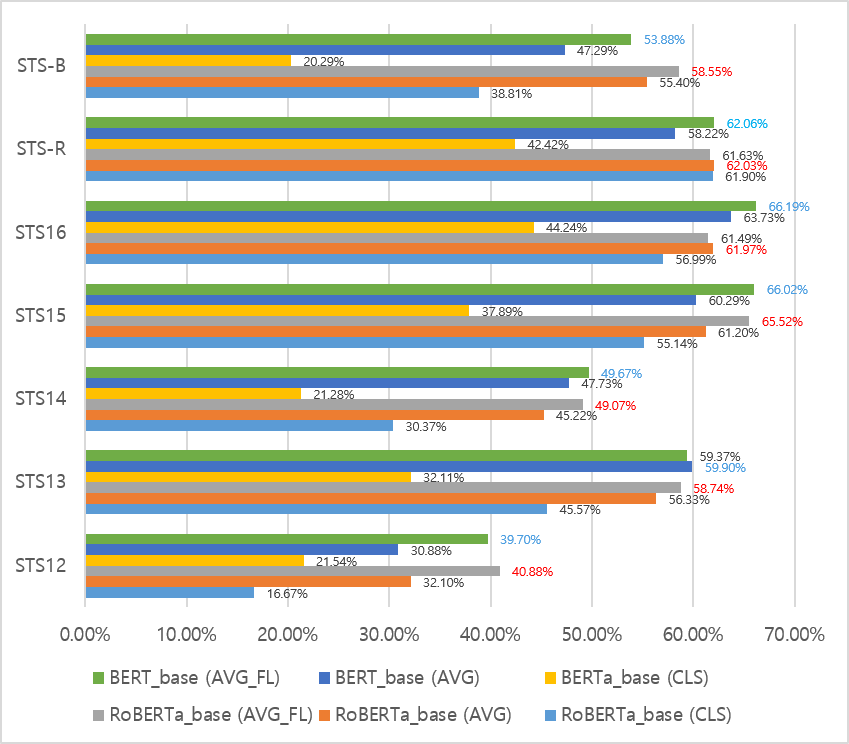}
\caption{Spearsman’s correlation score depending on the pooling methods of PLMs for each domain}
\label{fig:intro_fig2}

\end{figure}

Figure \ref{fig:intro_fig1} and \ref{fig:intro_fig2} show the Spearman's correlation score of each layer or pooling method in PLMs. We evaluated the test set of the standard semantic textual similarity (STS) dataset\cite{cer2017semeval, agirre2012semeval,agirre2013sem,agirre2014semeval,agirre2015semeval,agirre2016semeval,marelli2014sick}. 

The comparison of performance when pooling each layer shown in Figure \ref{fig:intro_fig1} indicates that using only a specific layer for pooling is insufficient. Other layers other than the last layer may contain substantial information for sentence representation. For example, for the STS benchmark (STS-B) task\citep{cer2017semeval}, BERT$_{base}$ with $[CLS]$ embedding scored the highest at the fourth layer (48.66\%), which is about 20\% higher than the last layer. 

Figure \ref{fig:intro_fig2} shows that simply pooling from more layers impedes the performance by comparing models pooled from the first and last layer and the last layer. In addition, there is no consistent tendency to compare effectiveness for a given layer between $[CLS]$ pooling and average pooling. 

Motivated by this point, we designed the attention networks and task-agnostic pooling methods to assign more weights to spots that need more focus in the layer and lead to representation vector optimization. Our proposed method outperforms previously fixed pooling strategies in contrastive learning. In addition, contrastive learning models with layer-wise attention pooling show a higher semantic search performance with the same parameters.

In summary, the contributions of this paper are as follows:
\begin{itemize}
\item We proposed layer-wise attention pooling\footnote{https://github.com/nlpods/LayerAttPooler} to assign weights to each layer and learn sentence representation fitted to a given task.
\item To our knowledge, our pooling strategy shows the best performance out of all InfoNCE-based loss functions for the sentence embedding tasks.
\item For the semantic search evaluation, we excluded the proposed pooling method in the inference phase and obtained better performance.
\end{itemize}

\section{Method}
In this section, we present a layer-wise pooling strategy based on attention mechanisms to improve the quality of sentence representations from language models. In addition, we describe the process of applying the proposed pooling strategy to be leveraged on three contrastive learning schemes.

\subsection{Layer-Wise Attention Pooling}
This paper proposes a new layer-wise pooling based on a multiplicative attention mechanism \cite{luong2015effective}. As shown in Figure \ref{fig:intro_fig1}, the performance with $[CLS]$ pooling varied dramatically according to which layer to pool from. There is no significant performance gap between layers when using $AVG$ pooling. It can be explained that each layer can contain different information for sentence representation, while average pooling can mitigate the information gap between layers.  

In Equation \ref{eq1}, $h^{a}$ is the $AVG$ representation, which is the mean vector of tokens in the sentence, and $h^{c}$ is the input representation $[CLS]$ of each layer. $\alpha _{i}$ is the importance of the $i$-th layer. In Equation \ref{eq2}, $h^{l}$ is the representation with the importance score per layer. In Equation \ref{eq3}, $h^{L}$ is the mean vector of $h^{l}$ and is the representation that contains the relevance of all layers (N is the number of layers). $W_{k}, W_{q}$ and $W_{v}$ are learnable parameters. 
\begin{equation} \label{eq1}\alpha _{i} = \frac{W_{q}h_{i}^{c}W_{k}h_{i}^{a}}{\sum_{j\in N}W_{q}h_{i}^{c}W_{k}h_{j}^{a}}\end{equation}

\begin{equation} \label{eq2}h_{i}^{l}=\sum_{j\in N}\alpha_{j}W_{v}h_{j}^{a}\end{equation}

\begin{equation} \label{eq3}h^{L}=\frac{1}{N}\sum_{i}^{N}h_{i}^{l}\end{equation}

We add a Multi-Layer Perceptron (MLP) layer randomly initialized after pooling, following the method in the \citet{gao2021simcse}, and keep it with random initialization. As for Equation \ref{eq4}, $h_{last}^{c}$ is the input representation $[CLS]$ of the last layer. We concatenate the input representation $h_{last}^{c}$ with the layer representation $h^{L}$ as the input of an MLP. Finally, $h$ is represented in the same dimension as the sentence representation dimension of the original language model through the MLP layer.

\begin{equation} \label{eq4}h^{CL}= [h_{last}^{c};h^{L}]\end{equation}

\begin{equation} \label{eq5}h = MLP(h^{CL})\end{equation}

\subsection{Contrastive Learning with Layer-wise Attention Pooling}
We prove that the proposed pooling strategy is effective with three training objectives $l_{i}$. 
\paragraph{Basic Supervised Contrastive Learning}
We use the basic supervised contrastive learning model proposed by \citet{chen2020simple}. This model learns the premise($x_{i}$) and entailment($x_{i}^{+}$) of the NLI(SNLI\cite{bowman2015large} + MNLI\cite{williams2018broad}) datasets. When $D={(x_{i}, x_{i}^{+})}_{i=1}^{m}$ is a set of paired samples, where $x_{i}$ and $ x_{i}^{+}$ are semantically related. And, it takes the cross-entropy objective with an in-batch negative\cite{chen2017sampling, henderson2017efficient}. $h_{i}$ and $h_{i}^{+}$ are representations of $x_{i}$ and $x_{i}^{+}$ through proposed pooling strategy. the training objective $l_{i}$ is : 

\begin{equation} \label{eq6} l_{i}=-log\frac{e^{sim(h_{i},h_{i}^{+})/\tau}}{\sum_{j=1}^{M}e^{sim(h_{i},h_{j}^{+})/\tau}}\end{equation}

$M$ is the mini-batch, and $\tau$ is the temperature hyperparameter and $sim(\cdot, \cdot)$ is the cosine similarity. 

\paragraph{Unsupervised Contrastive Learning}
Unsupervised contrastive learning uses $x_{i}^{+} = x_{i}$ in the collection of sentences $\{x_{i}\}_{i=1}^{m}$.  The idea is to use an independently sampled dropout mask for $x_{i}$ and $x_{i}^{+}$ which gets this to work as identical positive pairs during training. And, unsupervised contrastive learning denotes $h_{i}^{z} = f_{θ}(x_{i} , z)$ using $h$ obtained in Equation \ref{eq5}. $z$ is a random mask for dropout. It gets two embeddings with different dropout masks $z$, $z^{'}$ from the encoder with the same input twice, and the training objective $l_{i}$ is represented:

\begin{equation} \label{eq7}l_{i}=-log\frac{e^{sim(h_{i}^{z_{i}},h_{i}^{z_{i}^{'}})/\tau}}{\sum_{j=1}^{M}e^{sim(h_{i}^{z_{i}},h_{j}^{z_{j}^{'}})/\tau}}\end{equation}

In Equation \ref{eq7}, $z$ is the standard dropout of the transformer.

\paragraph{Supervised Contrastive Learning with Hard Negative}
Supervised contrastive learning with hard negative trains natural language inference (NLI) datasets. The NLI datasets are labeled, given one premise, as true(entailment), neutral, and definitely false (contradiction). The model predicts whether the relationship between two sentences is entailment, neutral, or contradiction. The positive pairs $(x_{i} , x_{i}^{+})$ use the entailment of the NLI(SNLI + MNLI) datasets. Next, contradiction pairs $(x_{i} , x_{i}^{-})$ from the NLI datasets are used as hard negatives. Thus, it expands from $(x_{i}, x_{i}^{+})$ to $(x_{i}, x_{i}^{+}, x_{i}^{-})$. And, $(x_{i}, x_{i}^{+}, x_{i}^{-})$ is represented as $(h_{i}, h_{i}^{+}, h_{i}^{-})$ through Equation \ref{eq5}. As a result, in Equation \ref{eq8}, the training objective $l_{i}$ is :

\begin{equation} \label{eq8} -log\frac{e^{sim(h_{i},h_{i}^{+})/\tau}}{\sum_{j=1}^{M}(e^{sim(h_{i},h_{j}^{+})/\tau}+e^{sim(h_{i},h_{j}^{-})/\tau})}\end{equation}

\section{Experiments}
\subsection{Experimental Setup}
Our main experiments uses the STS\cite{cer2017semeval, agirre2012semeval,agirre2013sem,agirre2014semeval,agirre2015semeval,agirre2016semeval,marelli2014sick} dataset. This data set consists of sentence pairs labeled with a similarity score between 0 and 5. The evaluation is done by the SentEval toolkit. The parameter setting of the model used in the experiment is written in Table \ref{table:PM} of the Appendix. Additionally, to measure the search effect and efficiency of the proposed model, it is evaluated on the same parameters as the original language model. We evaluate the performance of the semantic search\footnote{https://github.com/autoliuweijie/BERT-whitening-pytorch} with FAISS\footnote{https://github.com/facebookresearch/faiss} using the Quora Duplicate Questions Dataset\cite{shankar2021first} containing more than 400,000 pairs of questions.

\subsection{Main Results}
In Table \ref{table:STS}, we investigate whether the proposed layer-wise attention pooling of language models performs better in contrastive learning. The experiment compares performance by training on language models with three training objectives. All results evaluate sentence embeddings on all STS tasks. Equation \ref{eq6} is basic supervised learning proposed by \cite{chen2020simple}. And, Equations \ref{eq7} and \ref{eq8} are unsupervised, supervised learning proposed by \citet{gao2021simcse}. However, in this paper, we could not experiment with the same parameters due to hardware. Therefore, as specified in Table \ref{table:PM} of the Appendix, there is a difference from the original performance because it learns by choosing a low mini-batch size. $\dagger$ is the original performance, and $\ddagger$ is our reimplementations. As a result, the proposed pooling strategy shows higher performance in different language models and in all domains.

\begin{table*}[hbt!]
    \centering
     \resizebox{1.0\textwidth}{!}
    {
    \begin{tabular}{l|cccccccc}
    \hline
        \toprule
        \multicolumn{9}{c}{\textbf{Unsupervised Models}}\\ \hline\toprule
        \textbf{Model} & \textbf{STS12} & \textbf{STS13} & \textbf{STS14} & \textbf{STS15} & \textbf{STS16} & \textbf{STS-B} & \textbf{SICK-R} & \textbf{Avg} \\ \hline\hline
        BERT$_{base}$(CLS$_{Last}$)(Equation \ref{eq7})$\dagger$ & 68.40 & 82.41 & 74.38 & 80.91 & 78.56 & 76.85 & 72.23 & 76.25 \\
        RoBERTa$_{base}$(CLS$_{Last}$)(Equation \ref{eq7})$\dagger$ & 70.16 & 81.77 & 73.24 & 81.36 & 80.65 & 80.22 & 68.56 & 76.57 \\
        RoBERTa$_{large}$(CLS$_{Last}$)(Equation \ref{eq7})$\dagger$ & 72.86 & 83.99 & 75.62 & 84.77 & 81.80 & 81.98 & 71.23 & 78.89 \\ 
        \hline
        \multicolumn{9}{c}{\textbf{Our Reimplementations}}\\ \hline
        BERT$_{base}$(CLS$_{Last}$)(Equation \ref{eq7})$\ddagger$ & 69.53 & 78.98 & 75.50 & 80.07 & 79.01 & 78.28 & 71.35 & 76.10\\
        w/ LayerAttPooler(CLS$_{All}$ + AVG$_{All}$ attention) + (CLS$_{Last}$ concat) & \textbf{70.27} & \textbf{80.22} & \textbf{75.65} & \textbf{80.71} & \textbf{79.74} & \textbf{79.51} & \textbf{72.18} & \textbf{76.90}\\\hline
        RoBERTa$_{base}$(CLS$_{Last}$)(Equation \ref{eq7})$\ddagger$ & 68.72 & 78.29 & 74.35 & 80.40 & 80.83 & 80.14 & 68.71 & 75.92\\
        w/ LayerAttPooler(CLS$_{All}$ + AVG$_{All}$ attention) + (CLS$_{Last}$ concat) & \textbf{68.96} & \textbf{78.83} & \textbf{75.37} & \textbf{81.05} & \textbf{81.53} & \textbf{80.99} & \textbf{69.03} & \textbf{76.54} \\ \hline
        RoBERTa$_{large}$(CLS$_{Last}$)(Equation \ref{eq7})$\ddagger$ & 70.82 & 79.66 & 76.26 & 83.25 & 81.86 & 81.25 & 71.09 & 77.74\\
        w/ LayerAttPooler(CLS$_{All}$ + AVG$_{All}$ attention) + (CLS$_{Last}$ concat) & \textbf{71.52} & \textbf{79.86} & \textbf{76.86} & \textbf{83.50} & \textbf{82.38} & \textbf{84.56} & \textbf{71.46} & \textbf{78.59}\\
        \hline\toprule 
        \multicolumn{9}{c}{\textbf{Supervised Models}}\\ \hline\toprule
        BERT$_{base}$(CLS$_{Last}$)(Equation \ref{eq8})$\dagger$ & 75.30 & 84.67 & 80.19 & 85.40 & 80.82 & 84.25 & 80.39 & 81.57 \\ 
        RoBERTa$_{base}$(CLS$_{Last}$)(Equation \ref{eq8})$\dagger$ & 76.53 & 85.21 & 80.95 & 86.03 & 82.57 & 85.83 & 80.50 & 82.52 \\ 
        RoBERTa$_{large}$(CLS$_{Last}$)(Equation \ref{eq8})$\dagger$ & 77.46 & 87.27 & 82.36 & 86.66 & 83.93 & 86.70 & 81.95 & 83.76\\ 
        \hline
        \multicolumn{9}{c}{\textbf{Our Reimplementations}}\\ \hline
        BERT$_{base}$(CLS$_{Last}$)(Equation \ref{eq8})$\ddagger$ & 70.50 & 80.77 & 79.52 & 83.82 & 81.17 & 84.34 & 79.04 & 79.88\\ 
        w/ LayerAttPooler(CLS$_{All}$ + AVG$_{All}$ attention) + (CLS$_{Last}$ concat) & \textbf{71.34} & \textbf{80.84} & \textbf{79.76} & \textbf{83.86} & \textbf{81.42} & \textbf{86.76} & \textbf{79.80} & \textbf{80.54}\\ \hline
        RoBERTa$_{base}$(CLS$_{Last}$)(Equation \ref{eq8})$\ddagger$ & 70.80 & 81.31 & 79.60 & 83.48 & 82.86 & 85.71 & 79.77 & 80.50\\ 
        w/ LayerAttPooler(CLS$_{All}$ + AVG$_{All}$ attention) + (CLS$_{Last}$ concat) & \textbf{71.35} & \textbf{81.44} & \textbf{79.82} & \textbf{83.79} & \textbf{83.89} & \textbf{87.42} & \textbf{80.11} & \textbf{81.12}\\ \hline
        RoBERTa$_{large}$(CLS$_{Last}$)(Equation \ref{eq8})$\ddagger$ & 72.36 & 83.06 & 81.99 & 85.39 & 85.51 & 87.11 & 80.46 & 82.27\\ 
        w/ LayerAttPooler(CLS$_{All}$ + AVG$_{All}$ attention) + (CLS$_{Last}$ concat) & \textbf{72.65} & \textbf{84.41} & \textbf{82.31} & \textbf{86.38} & \textbf{85.54} & \textbf{87.58} & \textbf{81.56} & \textbf{82.92} \\
        \hline\toprule
        BERT$_{base}$(CLS$_{Last}$)(Equation \ref{eq6}) & 69.29 & 78.69 & 76.45 & 80.87 & 79.82 & 79.41 & 76.41 & 77.28 \\
        w/ LayerAttPooler(CLS$_{All}$ + AVG$_{All}$ attention) + (CLS$_{Last}$ concat) & \textbf{69.58} & \textbf{78.84} & \textbf{76.70} & \textbf{81.13} & \textbf{80.11} & \textbf{87.23} & \textbf{76.45} & \textbf{78.58} \\ \hline
        RoBERTa$_{base}$(CLS$_{Last}$)(Equation \ref{eq6}) & 68.85 & 77.28 & 74.67 & 80.11 & 80.80 & 87.42 & 76.51 & 77.95 \\
        w/ LayerAttPooler(CLS$_{All}$ + AVG$_{All}$ attention) + (CLS$_{Last}$ concat) & \textbf{69.51} & \textbf{78.72} & \textbf{75.97} & \textbf{81.32} & \textbf{81.47} & \textbf{89.58} & \textbf{76.83} & \textbf{79.06} \\ \hline
        RoBERTa$_{large}$(CLS$_{Last}$)(Equation \ref{eq6}) & 70.82 & 80.33 & 77.79 & 82.03 & 83.04 & 85.38 & 76.84 & 79.46 \\
        w/ LayerAttPooler(CLS$_{All}$ + AVG$_{All}$ attention) + (CLS$_{Last}$ concat) & \textbf{71.15} & \textbf{81.45} & \textbf{78.04} & \textbf{83.03} & \textbf{83.09} & \textbf{88.88} & \textbf{77.39} & \textbf{80.43} \\ 
        \hline
        \toprule
    \end{tabular}
    }
\caption{Performance of sentence embedding on all STS tasks (Spearman’s correlation). $\dagger$: published in \citet{gao2021simcse}; and $\ddagger$: models from our reimplementations. We are shown in bold the highest performance among models from our reimplementation.}
\label{table:STS}
\end{table*}

\subsection{Ablation Studies}
We investigate performance differences according to different pooling strategies in supervised contrastive learning. All results are reported in this section using the STS-B test set. All models extract sentence embeddings by adding an MLP layer as suggested in  \citet{gao2021simcse}. Table \ref{table:AS} shows the performance difference between the fixed pooling method and the layer-wise attention pooling. Additionally, we compare the representation concatenated between fixed pooling because we construct the $h$ representation by concatenating $h_{last}^{c}$ and $h^{L}$. The layer-wise attention pooling shows the results of ablation studies with $[CLS]$ and $AVG$. For $[CLS]_{All}$ and $AVG_{All}$, $h^{l}$ computes the importance between each layer and the others. In addition, $[CLS]_{All}$ + $AVG_{All}$ represent $h^{l}$ by calculating the importance between $[CLS]$ and $AVG$ of all layers. All of these methods show higher performance than the fixed pooling strategy. However, as described in Section 2, the pooling strategy concatenated with $[CLS]_{Last}$ shows the highest performance.

\begin{table}[hbt!]
    \centering
    \resizebox{0.5\textwidth}{!}
    {
    \begin{tabular}{l|c}
    \hline
        \toprule
        \textbf{Model} & \textbf{STS-B}\\ \hline\hline
        BERT$_{base}$(Equation \ref{eq8}) & \\
        w/ (CLS$_{Last}$) & 84.34 \\ 
        w/ (AVG$_{Last}$) & 84.84 \\ 
        w/ (AVG$_{FL}$) & 84.76 \\ 
        w/ (AVG$_{Last}$+AVG$_{FL}$ concat) & 84.93 \\ 
        w/ (CLS$_{Last}$+AVG$_{Last}$ concat) & 85.11 \\ 
        w/ LayerAttPooler(CLS$_{All}$ attention) & 85.45 \\ 
        w/ LayerAttPooler(AVG$_{All}$ attention) & 85.72 \\ 
        w/ LayerAttPooler(CLS$_{All}$ + AVG$_{All}$ attention) & 86.57 \\
        w/ LayerAttPooler(CLS$_{All}$ + AVG$_{All}$ attention) + (CLS$_{Last}$ concat)  & \textbf{86.76} \\
        \hline
        \toprule 
    \end{tabular}
    }
\caption{Ablation studies of different pooling methods in supervised model on STS-B task (Spearman’s correlation)}
\label{table:AS}
\end{table}

\subsection{Semantic Search Results}
In Table \ref{table:SS}, we compare semantic search speed and performance on the same parameters. This experiment proves that the proposed pooling strategy is effective for training the language models and also for semantic search performance with the same parameters during inference. Sentence embeddings for all supervised learning models use $[CLS]_{Last}$. MRR@10 is used to measure the performance of semantic search, and Average Retrieval Time (ms) measures retrieval efficiency. Memory Usage (GB) shows memory usage. FAISS experiments in CPU mode. nlist = 1024 and the CPU is Intel(R) Xeon(R) Gold 6230R CPU @ 2.10GHz. Result shows that the performance of semantic search is higher when the proposed pooling strategy is used during training.

\begin{table}[hbt!]
    \centering
    \resizebox{0.5\textwidth}{!}
    {
    \begin{tabular}{l|c|c|c}
    \hline
     \toprule
        \multirow{2}{*}{\textbf{Model}} & \multirow{2}{*}{\textbf{MRR@10}} &  \textbf{Average Retrieval Time} &  \textbf{Memory Usage} \\
        & & \textbf{(ms)} & \textbf{(GB)}  \\  \hline\hline
        BERT$_{base}$(Equation \ref{eq8}) & & & \\
        w/ (CLS$_{Last}$) & 63.48 & 1.46 & 0.25 \\ 
        w/ LayerAttPooler (train) & \textbf{64.32} & \textbf{1.45} & 0.25 \\\hline
        RoBERTa$_{base}$(Equation \ref{eq8}) & & & \\
        w/ (CLS$_{Last}$) & 63.89 & 1.56 & 0.25 \\ 
        w/ LayerAttPooler (train) & \textbf{65.05} & \textbf{1.48} & 0.25 \\\hline
        RoBERTa$_{large}$(Equation \ref{eq8}) & & & \\
        w/ (CLS$_{Last}$) & 65.85 & 2.22 & 0.33 \\ 
        w/ LayerAttPooler (train) & \textbf{66.32} & \textbf{2.21} & 0.33 \\
    \hline
     \toprule
    \end{tabular}
    }
\caption{Performance of semantic search evaluation using the Quora Duplicate Questions Dataset with FAISS. w/ LayerAttPooler (train) : remove layer-wise attention pooling after training}
\label{table:SS}
\end{table}

\section{Conclusion}
In this work, we propose layer-wise attention pooling to capture the importance of the weight in each layer for the pre-trained language models (PLMs). Training layer-wise attention layer with contrastive learning objectives outperforms BERT and variants of PLMs. No matter what pooling method is used, our model achieved higher scores than prior state-of-the-art models. In addition, this layer-wise attention technique also can be exploited in semantic search tasks, in which more cost-efficient computation (i.e. less latency and memory usage) is required. The model trained with our method obtained higher performance with the same or less time and memory usage, even if the added attention layer is detached in the inference stage. 

\section*{Acknowledgements}
This work was supported by Institute of Information \& communications Technology Planning \& Evaluation(IITP) grant funded by the Korea government(MSIT) (No. 2020-0-00368, A Neural-Symbolic Model for Knowledge Acquisition and Inference Techniques) and by the MSIT(Ministry of Science and ICT), Korea, under the ITRC(Information Technology Research Center) support program(IITP-2022-2018-0-01405) supervised by the IITP(Institute for Information \& Communications Technology Planning \& Evaluation) In addition, This research was supported by Basic Science Research Program through the National Research Foundation of Korea(NRF) funded by the Ministry of Education(NRF-2021R1A6A1A03045425)

\bibliography{acl}
\bibliographystyle{acl_natbib}

\newpage
\appendix
\section{Training Details}

\begin{table}[hbt!]
    \centering
     \resizebox{0.5\textwidth}{!}
    {
    \begin{tabular}{l|cc}
    \hline
        \toprule
        \multicolumn{3}{c}{\textbf{Unsupervised Models}}\\ \hline\toprule
        \textbf{Models} & \textbf{Batch Size} & \textbf{Learning Rate} \\ \hline\hline
        BERT$_{base}$(Equation \ref{eq7}) & 64 & 3e-5 \\
        w/ LayerAttPooler & 64 & 3e-5 \\
        RoBERTa$_{base}$(Equation \ref{eq7}) & 256 & 1e-5 \\
        w/ LayerAttPooler & 256 & 1e-5 \\
        RoBERTa$_{large}$(Equation \ref{eq7}) & 256 & 3e-5 \\ 
        w/ LayerAttPooler & 256 & 3e-5 \\
        BERT$_{base}$(DiffCSE) & 64 & 7e-6 \\
        w/ LayerAttPooler & 64 & 3e-5 \\
        \hline
        \toprule
        \multicolumn{3}{c}{\textbf{Supervised Models}}\\ \hline\toprule
        BERT$_{base}$(Equation \ref{eq6}) & 256 & 5e-5 \\
        w/ LayerAttPooler & 256 & 1e-5 \\
        RoBERTa$_{base}$(Equation \ref{eq6}) & 256 & 5e-5 \\
        w/ LayerAttPooler & 256 & 3e-5 \\
        RoBERTa$_{large}$(Equation \ref{eq6}) & 256 & 1e-5 \\ 
        w/ LayerAttPooler & 256 & 5e-5 \\
        \hline\toprule
         BERT$_{base}$(Equation \ref{eq8}) & 256 & 5e-5 \\
        w/ LayerAttPooler & 256 & 2e-5 \\
        RoBERTa$_{base}$(Equation \ref{eq8}) & 256 & 5e-5 \\
        w/ LayerAttPooler & 256 & 3e-5 \\
        RoBERTa$_{large}$(Equation \ref{eq8}) & 256 & 1e-5 \\ 
        w/ LayerAttPooler & 256 & 5e-5 \\
        \hline
        \toprule
    \end{tabular}
    }
\caption{Batch sizes and learning rate for each models}
\label{table:PM}
\end{table}

Due to hardware problems, Equations \ref{eq7} and \ref{eq8} train at a smaller batch size than the \citet{gao2021simcse} paper. The GPU used in the experiment is RTX 8000, and the the hyperparameters are specified in the Table \ref{table:PM}.

\section{Experiments on Different Model}

\begin{table}[hbt!]
    \centering
     \resizebox{0.5\textwidth}{!}
    {

    \begin{tabular}{l|c}
    \hline
        \toprule
        \multicolumn{2}{c}{\textbf{DiffCSE model}}\\ \hline\toprule
        \textbf{Model} & \textbf{STS-B}\\ \hline\hline
        BERT$_{base}$(CLS$_{Last}$) (w/o BatchNorm)$\dagger$ & 83.23 \\
        w/ LayerAttPooler(CLS$_{All}$ + AVG$_{All}$ attention) + (CLS$_{Last}$ concat) & \textbf{83.87} \\
        \hline
        \toprule
    \end{tabular}
    }
\caption{Development set results of STS-B. $\dagger$: published in \citet{chuang2022diffcse}; Bold shows the highest performance among models from our reimplementation.}
\label{table:DiffCSE}
\end{table}

We experiment with whether the proposed pooling strategy is effective for a contrastive learning model with a different structure. DiffCSE model \cite{chuang2022diffcse} improves the performance of sentence representation by adding generator and discriminator structures of ELECTRA \cite{clark2020electra}.  While training, DiffCSE freezes the generator’s weight and updates the sentence encoder and discriminator for sentence embedding with the contrastive learning objective. However, the discriminator is not used for inference since only representations from the sentence encoder and generator are needed. We applied our proposed pooling strategy to the sentence encoder with a contrastive learning objective. As a result, layer-wise attention pooling improves the performance of the DiffCSE model (Table \ref{table:DiffCSE}). We use the one linear layer with the tanh activation function following SimCSE as in Equation \ref{eq5}, while DiffCSE uses a two-layer pooler with Batch Normalization (BatchNorm) \cite{ioffe2015batch}. However, BatchNorm is not used for a fair comparison of results.

\section{Analysis Attention Weights over Layers}

\begin{figure}[H]
\centering

\subfloat[Attention scores of LayerAttPooler(CLS$_{All}$ + AVG$_{All}$ \\attention) + (CLS$_{Last}$ concat)]{
	\label{subfig:concat}
	\includegraphics[width=0.5\textwidth]{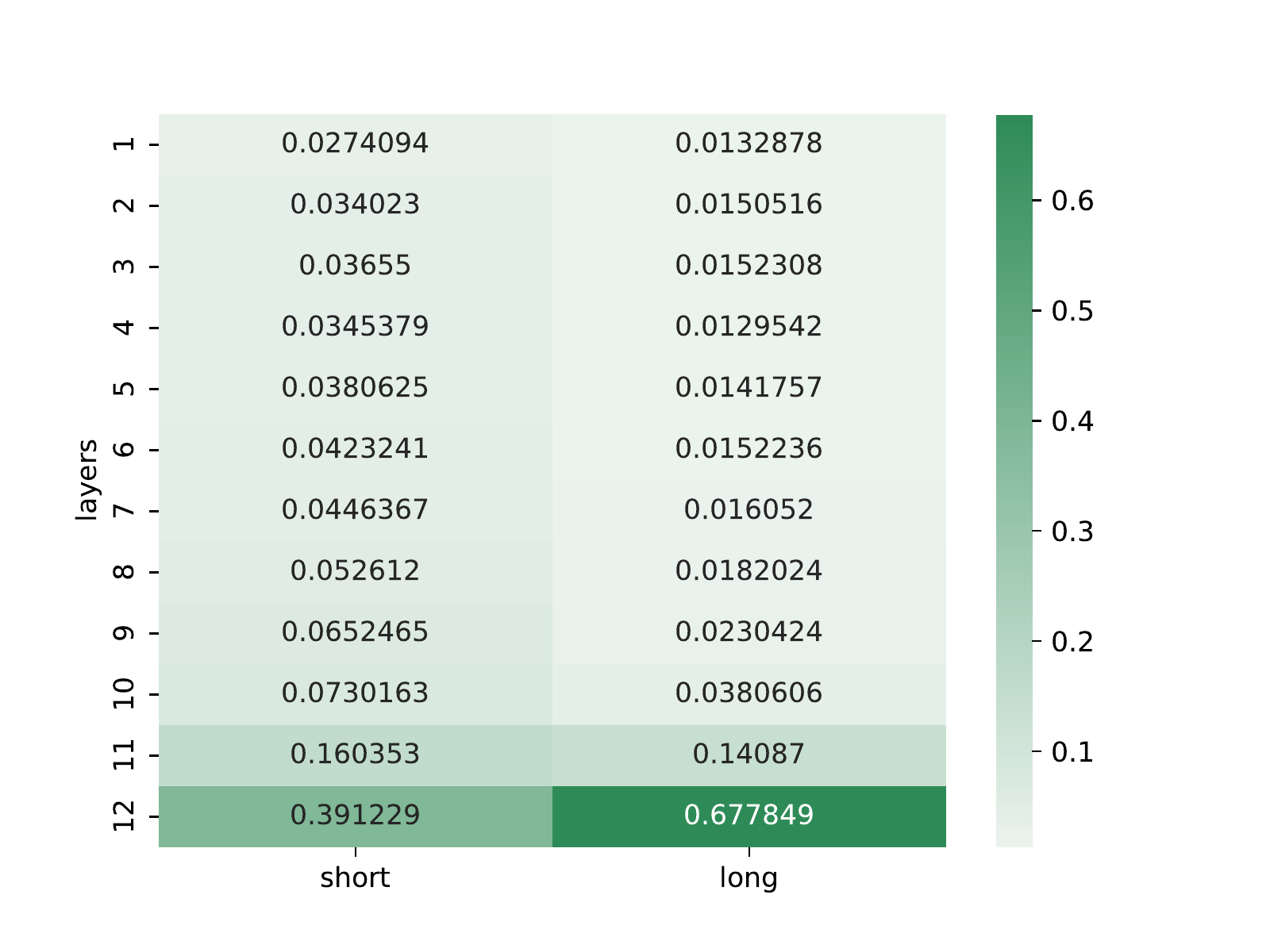} } 

\subfloat[Attention scores of LayerAttPooler(CLS$_{All}$ + AVG$_{All}$ \\attention)]{
	\label{subfig:layer}
	\includegraphics[width=0.5\textwidth]{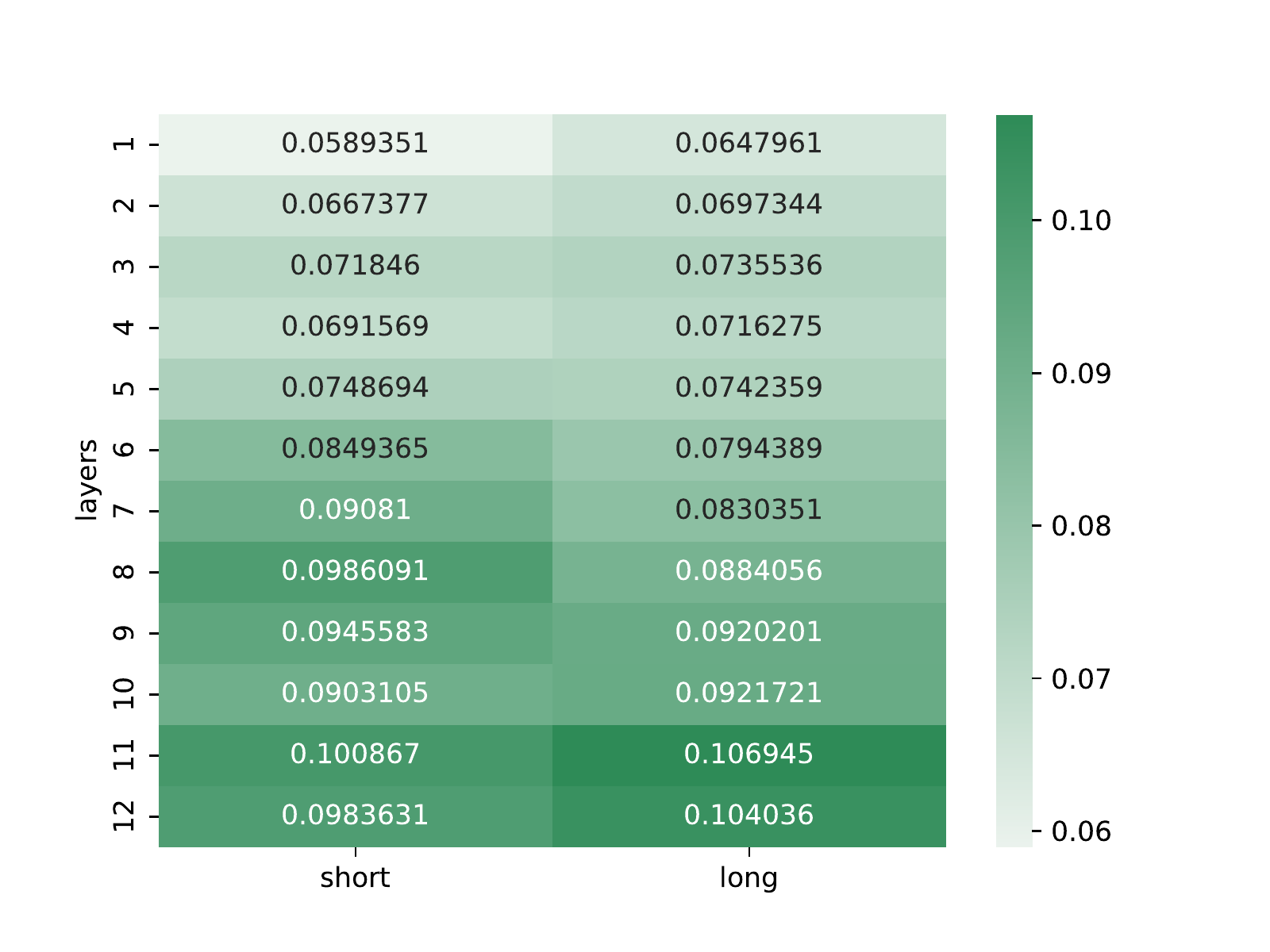} } 
	
\caption{Attention scores of layer-wise pooling only (b) and concatenating the $[CLS]_{Last}$ representation of the last layer (a) on a sentence "You should do it." (short) and "People on motorcycles wearing racing gear ride around a racetrack." (long) sentences. These scores are implemented on BERT$_{base}$.}
\label{fig3}
\end{figure}

We also analyze the layer-wise attention scores depending on the length of sentences. Figure 3 (a) case explains that the last layer relatively contains more information than other layers by the $[CLS]$ token of the last layer. However, the attention score of the last layer is calculated differently for the long and short sentences. Figure 3 (b) case indicates that other layers than the last layer have substantial information for the same sentence, and the balanced attention weight per layer supports it.

\end{document}